%% file: acl_latex.tex
\pdfoutput=1

\documentclass[11pt]{article}
\usepackage{authblk}
\usepackage{longtable}

\usepackage{acl}

\usepackage{times}
\usepackage{latexsym}
\usepackage{amsmath}
\usepackage{svg}

\usepackage{graphicx}
\usepackage{subcaption}
\usepackage{makecell}
\usepackage{multirow}
\usepackage{algorithm}
\usepackage{algorithmic}
\usepackage{amsmath}
\usepackage{amsfonts}
\usepackage{booktabs}
\usepackage{siunitx}
\usepackage{float} 
\usepackage{enumitem} 
\usepackage{booktabs}
\usepackage{listings}
\usepackage{soul}
\usepackage{xcolor}
\usepackage{enumitem}
\usepackage{tabularx}

\usepackage[T1]{fontenc}

\usepackage[utf8]{inputenc}

\usepackage{microtype}

\newcommand{\SystemName}{VeraCT Scan}

\NewDocumentCommand{\shizhe}
{ mO{} }{\textcolor{blue}{\textsuperscript{\textit{shizhe}}\textsf{\textbf{\small[#1]}}}}

\title{\SystemName: Retrieval-Augmented Fake News Detection with Justifiable Reasoning}

\author[1]{\textbf{Cheng Niu}}
\author[1]{\textbf{Yang Guan}}
\author[1]{\textbf{Yuanhao Wu}}
\author[1]{\textbf{Juno Zhu}}
\author[1]{\textbf{Juntong Song}}
\author[1]{\textbf{Randy Zhong}}
\author[1]{\\\textbf{Kaihua Zhu}}
\author[1]{\textbf{Siliang Xu}}
\author[2]{\textbf{Shizhe Diao}}
\author[3]{\textbf{Tong Zhang}}
\affil[1]{NewsBreak}
\affil[2]{Hong Kong University of Science and Technology}
\affil[3]{University of Illinois Urbana-Champaign}
\affil[ ]{ {cheng.niu@newsbreak.com}}

\begin{document}
\maketitle
\begin{abstract}
The proliferation of fake news poses a significant threat not only by disseminating misleading information but also by undermining the very foundations of democracy. 
The recent advance of generative artificial intelligence has further exacerbated the challenge of distinguishing genuine news from fabricated stories. 
In response to this challenge, we introduce \SystemName, a novel retrieval-augmented system for fake news detection. This system operates by extracting the core facts from a given piece of news and subsequently conducting an internet-wide search to identify corroborating or conflicting reports. Then sources' credibility is leveraged for information verification. 
Besides determining the veracity of news, we also provide transparent evidence and reasoning to support its conclusions, resulting in the interpretability and trust in the results. In addition to GPT-4 Turbo, Llama-2 13B is also fine-tuned for news content understanding, information verification, and reasoning. Both implementations have demonstrated state-of-the-art accuracy in the realm of fake news detection\footnote{Our demo is available at \url{https://veractscan.newsbreak.com/}. Demo video at \url{https://youtu.be/t1__iuOG9H8}.}. 

\end{abstract}

\section{Introduction}
The contemporary digital landscape is rife with the proliferation of fake news, presenting a multifaceted challenge that undermines public discourse, affects democratic processes, and incites real-world consequences~\citep{vasu2018fake}. 
Fake news, characterized by the deliberate dissemination of misinformation, exploits the rapid spread of information online, often outpacing the verification processes that traditional media outlets adhere to. 

Fake news detection is defined as the process of identifying and verifying the veracity of news content, employing various computational and manual methods. 
This process involves distinguishing between true and false information, considering the intent behind the information dissemination, whether it be to mislead, harm, or manipulate public opinion.

Traditional approaches in fake news detection have primarily focused on the linguistic features,
also called content-based detection~\citep{castillo2011information, perez2017automatic, giachanou2019leveraging, przybyla2020capturing, giachanou2020multimodal, sheikhi2021effective, kirchknopf2021multimodal, zhou2020fakenewsearly}, which demands laborious feature engineering and is ineffective when the fake news is written by imitating the real news to mislead intentionally.
Another line of research is the social context-based method~\citep{qazvinian2011rumor, baly2018predicting, shu2019role, monti2019fake, nan2023exploiting}, which analyzes the interactions among users, publishers, and posts.
However, the feasibility of obtaining user information is challenging for the real-world application.
A more recent research approach is the knowledge-based method~\citep{hu2021compare, saeed2022crowdsourced, pan2023fact, chen-etal-2023-causal, 10.1145/3580305.3599873,zhang2023llmbased,li2023selfchecker}, which discerns the veracity of a factual claim by comparing against the evidence retrieved from external knowledge base.
However, current approaches often do not fully utilize external resources like the Internet. 
Additionally, there is a lack of development and optimization of a comprehensive end-to-end pipeline that includes news comprehension, search optimization, verification, and reasoning.


In this paper, we introduce \SystemName, a novel retrieval-augmented system for fake news detection.
{\SystemName} initiates this process by identifying key factual claims across multiple levels of granularity. For each identified factual claim, a comprehensive internet search is conducted to gather relevant information. Then, the veracity of the news is determined by combining this typically disparate and conflicting information, taking into account the varying degrees of source credibility. 
To increase the trustworthiness of our approach, we underscore the necessity of a transparent reasoning process and provide rationales for each supporting or conflicting judgment.

In summary, our main contributions are:

\begin{enumerate}[label=(\roman*),nolistsep]
\item We introduce \SystemName, that operates across multiple levels of information granularity, employing optimized information retrieval techniques to enhance fake news detection performance.
\item We investigate the generation of verification rationales as a means to increase the system's transparency and trustworthiness. Additionally, we address the management of conflicting evidence by leveraging the credibility of sources, thereby improving the reliability of the verification process.
\item We conduct a comprehensive evaluation of {\SystemName} using several fake news detection datasets. 
Our results demonstrate that the system achieves state-of-the-art performance in news verification tasks, employing both prompted and fine-tuned LLMs.
\end{enumerate}

\section{Related Work}
In this section, we first review the progress of fake news detection and then discuss the retrieval-augmented generation methods.

\subsection{Fake News Detection}
Existing fake news detection methods can be categorized into three types: 
1) Content-Based Methods~\citep{sheikhi2021effective, perez2017automatic, castillo2011information, przybyla2020capturing, giachanou2019leveraging, huang-etal-2023-faking, giachanou2020multimodal, kirchknopf2021multimodal, nakamura-etal-2020-fakeddit, chen-etal-2023-causal,zhou2020fakenewsearly} which analyze articles' linguistic features (e.g., text length, punctuation usage, emotion symbols) to differentiate fake news from real ones. However, these methods demand laborious feature engineering and are often ineffective when fake news is written to intentionally mislead readers.
2) Social Context-Based Methods~\citep{shu2019role, nan2023exploiting, baly2018predicting, monti2019fake, qazvinian2011rumor} which analyze the interactions among users, publishers, and posts to detect fake news. However, the feasibility of obtaining user information in the news propagation process presents challenges for the real-world applicability of this method.
3) Fact-Based Methods~\citep{saeed2022crowdsourced, pan2023fact, hu2021compare, Xu2023CounterfactualDF, chen-etal-2023-causal,cheung2023factllama} which focus on factual claim verification by comparing against external knowledge. These methods fall short in providing an end-to-end solution that considers information seeking and the management of conflicting evidence.

Recently, 
~\citet{wang2023explainable} leverage large language models (LLMs) to decompose complex claims into sequences of first-order logic, and then guide the search and information verification. 
Different from their work, we propose a pipeline that includes full steps to classify fake news. ~\citet{10.1145/3580305.3599873} outlines a multi-step process for detecting fake news, which consists of news summarization, searching, and verification. 
In contrast to their method, we employ LLMs instead of specifically trained encoder-decoder transformers for these natural language processing tasks. 
In addition, we leverage source credibility to differentiate conflicting evidences, a common challenge in real-world news verification that has rarely been explored in previous research.

\subsection{Retrieval-Augmented Generation}

The integration of retrieval-augmented generation (RAG) allows LLMs to extend beyond the limits of the training corpus by retrieving information from external knowledge bases before the generative process~\citep{lewis2020retrieval, chen2023benchmarking}.
RAG has emerged as a solution to overcome the limitations of LLMs including the challenge of out-of-date knowledge and the tendency to produce hallucinations or irrelevant and factually incorrect content.
By integrating external, up-to-date documents into the generation process, LLMs can generate more reliable responses across a broad spectrum of tasks, including open-domain question answering~\citep{izacard2020leveraging, trivedi2022interleaving, li2022large, xu2024can}, dialogue systems~\citep{cai2018skeleton, peng2023check}, and code generation~\citep{zhou2022docprompting}.
RAG is also commonly integrated into commercial chatbot products to provide updated information, e.g Perplexity\footnote{\url{https://www.perplexity.com}} and Gemini\footnote{\url{https://gemini.google.com}}.
In this paper, we leverage RAG for fake news detection by generating both verdicts and justifications. 
\input{figures/system}

\section{Approach}
\label{sec:approach}

In this paper, the term "claim" refers to the fact stated in a news article. The terms "factual claim extraction" and "fact extraction" are used interchangeably throughout the paper.

Figure~\ref{fig:system} shows the main workflow of {\SystemName}.
We prompt GPT-4 Turbo for key fact extraction, query generation, verification, and rationale generation (See Appendix~\ref{sec:appendix-prompt} for prompts being used). These individual components can be easily exchange to other LLMs or search engines. In this work, the outputs from GPT-4 Turbo, supplemented with manual reviews, serve as training data to fine-tune Llama-2 13B~\cite{touvron2023llama}, enabling it to support these tasks as well. 
Regarding the search component, we employ both Google and our proprietary in-house news search engine for comprehensive information retrieval.

\subsection{Key Fact Extraction}
In this paper, 
we focus on identifying facts at two levels of granularity: (i) the primary fact reported by the news story and (ii) all the salient facts being reported in the news article. 

Given that the internet search operates as a stateless module, we instruct the LLM in the prompt to ensure each key fact is self-contained with its information. This approach allows the search function to generate queries for each key fact independently, without relying on additional context. 

In line with the previous research ~\citep{shahandashti2024evaluating}, our manual review has confirmed the high quality of key facts being identified by GPT-4 Turbo.

\subsection{Query Generation and Search}
When verifying a fact, we prompt GPT-4 Turbo to generate search queries. We allow up to three queries per fact to search the Internet. Subsequently, GPT-4 Turbo assesses the relevance of the results returned by each query. The goal is to identify the shortest sequence of queries that can retrieve all the relevant information. This optimal query sequence is then utilized to fine-tune Llama-2 13B, enabling its query generation capabilities.

We have developed a proprietary search engine designed to support news searches for articles published within the last six months. This search engine is especially effective in searching articles hosted on NewsBreak platform and can be used in NewsBreak APP. To ensure comprehensive search results, we also utilize the Google search API \footnote{\url{https://developers.google.com/custom-search/v1/overview}}.

\subsection{Fact Verification and Rationale Generation}
Once the search results are retrieved, each fact is evaluated against them. GPT-4 Turbo is prompted to iterate each of the search results, and determine whether the search result supports, conflicts with, or is unrelated to the fact. If the search result aligns with the fact, it is labeled as "support". If it contradicts the fact, it is labeled as "negate". If the fact is not mentioned or only partially mentioned in the search result, the label "baseless" is applied. Besides, a rationale is generated to justify the judgment. A concrete example of our pipeline is shown in Appendix~\ref{sec:appendix-sample}.

\subsection{Source Credibility and Final Decision}
\label{sec:classifier}
When researching a given topic, it is common to encounter conflicting information on the Internet. To avoid bias from single source, multiple sources are used to corroborate each other. Therefore, assessing the credibility of each information source is crucial. Mediabiasfactcheck.com is one of the most comprehensive resources for assessing media bias on the internet, offering credibility ratings for over 8,000 news publishers. Similarly, NewsBreak has developed a proprietary 5-level credibility rating system for more than 30,000 publishers. While NewsBreak's ratings are also based on the credibility of source content, unlike mediabiasfactcheck.com, NewsBreak does not identify the political bias of the sources. 

In this paper, NewsBreak's rating systems serves as features to train a LightGBM\cite{ke2017lightgbm} classifier that determines the likelihood of a fact claim being true. Besides, domain and verification flags (i.e. support, negate, or baseless) from each search result are also used as classification features.

\subsection{Llama-2 13B Fine Tuning}
To enhance service stability, response speed, and reduce costs, Llama-2 13B is fine-tuned to support our fake news detection pipeline.

\paragraph{Dataset} 
Following previous studies\cite{zhou2023lima, alpaca}, we utilize a mixed dataset of diverse tasks for supervised fine-tuning~(SFT). Outputs of GPT-4 Turbo from the tasks described above are used as part of the training data. Specifically, we purposely modify some key factual claims being extracted from news articles into fake ones when generating claim verification data set. Besides, the following datasets have also been incorporated into the training set:

\begin{enumerate}[nolistsep]
    \item QA with RAG: GPT-4 generated answers to questions in NewsBreak search logs using knowledge retrieved from our proprietary search engine.
    \item WebGLM\cite{liu2023webglm}: web-enhanced question-answering dataset.
    \item No robots\cite{no_robots}: a diverse instruction fine-tuning dataset created by skilled human annotators.
    
\end{enumerate}

The training data distribution is shown in Table~\ref{tab:dataset}. 
This design allows a single model to handle both general question-answering and specialized news verification tasks, resulting in significant reductions in inference costs.

\input{tables/dataset}

\paragraph{Hyper parameters}
To enhance the capability of processing long inputs, we trained the model with RoPE scaling\cite{su2023roformer, liu2023scaling}. Specifically, we adjusted the context window size in SFT to be twice as large as that in the original Llama-2 model, setting it to 8192 tokens, and we set the scaling factor at 2.0. We employed full training with an initial learning rate of 1e-5, and limited the training to 1 epoch. The training process was executed on four NVIDIA A100 GPUs.

\subsection{Key Task Evaluations} 
The end-to-end metrics will be present in Section~\ref{sec:result}. In this section, we present the performance metrics for the critical components.

With GPT-4 Turbo outputs as the gold standard, we benchmarked the finetuned Llama-2 model on key fact extraction, query generation, and rationale generation. ROUGE scores \cite{lin-2004-rouge} were employed as the metrics, as shown in Table 2. 

\input{tables/key_task}

For the fact verification accuracy, micro-F1 score was employed as the metric. According to human review, GPT-4 Turbo achieved a score of 0.805, while the finetuned Llama-2 model achieved 0.759.

\section{Experimental Settings}
In this section, we conduct comprehensive fake news detection benchmarks using multiple datasets.

\subsection{Datasets}

\paragraph{BuzzFeedNews\cite{buzzfeednews16}} 
This dataset consists of news articles shared on Facebook during the week surrounding the 2016 U.S. election. It includes data collected from nine different news agencies, spanning from September 19 to 23, and then September 26 and 27. Each article was fact-checked by a team of five BuzzFeed journalists. The articles are categorized under four labels: mostly true, mostly false, a mix of true and false, and no factual content. In line with \citet{shu2019beyond}, we utilize the subset of 182 news articles for our benchmark. Each article in this subset has been assigned one of two binary labels (true or fake news), making it suitable for our binary classification setting.

\paragraph{Fakenewsnet~\citep{shu2017fake,shu2017exploiting,shu2018fakenewsnet}} A fake news dataset characterized by its rich diversity, including news articles and social context. The contents have been sourced from PolitiFact\footnote{\url{https://www.politifact.com}} and GossipCop\footnote{\url{https://www.gossipcop.com} is now closed}, with most of them dating back to before 2018. In this paper, we have chosen to utilize the PolitiFact portion due to its high quality, as all the facts have been verified by domain experts.

\paragraph{LLMFake~\citep{chen2023llmgenerated}} A misinformation dataset is further modified by LLMs such as ChatGPT. These models utilize various techniques, including paraphrasing, rewriting, etc. for information manipulation. The information within this dataset traces back to 2020 or earlier.

\paragraph{PolitiFact-Snopes-2024} The dataset was manually collected from the prestigious fact-checking organizations PolitiFact and Snopes\footnote{\url{https://www.snopes.com}}. It includes approximately 1,200 verifiable claims along with the fact-check rating labels that determine the level of truthfulness for each claim. The clarifications for the labels and the additional detailed analysis reports were not collected. Non-text-based claims were filtered out, and exclusive fact-checks with supporting sources specific to these organizations were also filtered out.

\paragraph{FakeNews2024} This dataset consists 46 real news and 63 fake news articles. All the news articles are less than one year old, and are confirmed by NewsBreak moderation team.

The first three datasets were selected to enable a comparison of our system against three distinct fake news detection methods: content-based, LLMs-based, and retrieval-augmented approaches. The last two datasets are used to demonstrate our approach's ability to detect the latest fake news.

\subsection{Evaluation Metrics}
For the existing datasets, we strive to employ the same evaluation metrics that have been utilized in prior studies to enable direct comparisons.

For BuzzFeedNews, we report the precision, recall, and F1 scores related to fake news, as well as the accuracy for the entire dataset.
For Fakenewsnet, PolitiFact-Snopes-2024, and FakeNews2024, we report the precision (P-F), recall (R-F), and F1 score (F1-F) of the fake news, the precision (P-T), recall (R-T), and F1 score (F1-T) of the real news, as well as the Micro F1 score (F1) of the overall dataset.
For LLMFake, we report the detection success rate, which is calculated by the percentage of successfully identified fake news~\cite{chen2023llmgenerated}.

\subsection{Implementation Details}
To aid in the verification of news articles, the main factual claim of each news article is identified and then compared against internet search results. 
To ensure a fair comparison, we have developed heuristics to carefully filter out fact-checking content from search engine results in all the experiments below.

The datasets above except LLMFake are each aggregated to train the final LightGBM classifier, utilizing the features outlined in Section~\ref{sec:classifier}, and subsequently report the end-to-end accuracy. Both the training and testing processes are conducted using a 5-fold cross-validation approach. We also provided baseline benchmarks for comparison.

\section{Experimental Results}
\label{sec:result}

\input{tables/buzzfeed}

The performance with the BuzzfeedNews dataset is detailed in Table~\ref{tab:result_buzzfeed}. The baseline methods being reported in~\citet{zhou2020fakenewsearly} utilize features from article content, and outperform our approach. This outcome is expected since BuzzfeedNews dataset focuses primarily on a limited range of topics, specifically the 2016 US election. The nature of the fake news within this dataset allows it to be effectively modeled through content features. Furthermore, the fake news articles are approximately 7 years old, posing additional challenges for search engines in retrieving relevant evidences.

In Table~\ref{tab:result_fakenewsnet}, we present a performance comparison between {\SystemName} and another retrieval-augmented system, utilizing the FakeNewsNet dataset. Our two implementations, GPT-4 Turbo and the fine-tuned version of Llama-2 13B, both exhibit superior accuracy. This comparison underscores the efficacy of using either prompted or fine-tuned LLMs over specialized encoder-decoder transformers that have been specifically trained for this task.

\input{tables/fakenewsnet}

Table~\ref{tab:result_llm_generated} presents the detection performance using LLMFake. Notably, although the news articles in LLMFake are from 2020 or earlier—falling within GPT-4's inherent knowledge base, {\SystemName} significantly outperforms GPT-4 in verification accuracy. Notably, the Llama-2 13B implementation also wins 7 out of 12 benchmarks. This underscores the benefits and efficacy of incorporating knowledge from the Internet. It is important to note that LLMFake verification is not straightforward. According to~\citet{chen2023llmgenerated}, the accuracy of human annotations falls well below 40\%.

\input{tables/llm_generated}

In Tables~\ref{tab:result_poli_sno} and \ref{tab:result_latest_news}, we present the detection accuracy of our system when tested against the latest news articles. Unlike BuzzFeedNews, these two datasets consist of a wide variety of topics, including politics, entertainment, international warfare, and more. Both implementations of our system present relatively high detection accuracy, and underscores the effectiveness in verifying the latest news. Our approach benefits significantly from the enhanced efficiency of both Google and our proprietary search engine in sourcing relevant evidences for recent news.

\section{Conclusion and Future Work}
In this paper, we present {\SystemName}, a  novel retrieval-augmented system for fake news detection. Two of our implementations, properly prompted GPT-4 Turbo and fine-tuned Llama-2 13B demonstrated notable accuracy in detection. Specifically, the GPT-4 Turbo implementation exhibited state-of-the-art performance in several datasets. {\SystemName} is especially successful in identifying the latest instances of fake news. This emphasizes the critical role of search result relevance in gathering compelling evidence.

Our observations reveal that the rationales generated by LLMs offer rich insights into potentially dubious aspects with a high degree of details. As a future work, we plan to investigate the potential of using these rationales as input features for the final verification classifier. And throughout our evaluations, Llama-2 13B consistently lags behind GPT-4 Turbo in terms of detection accuracy. We will explore more effective fine-tuning strategies to narrow this performance gap.

Furthermore, we observe that within the entire system, the majority of errors occur during the verification stage, with a smaller fraction arising during the claim extraction phase. The causes of these errors include: (i) Irrelevant search results used for verification. (ii) Updated news events leading to outdated reports being used for verification. (iii) Each report only supporting a part of the claim, necessitating the proper merging of relevant information from multiple news reports for full verification. (iv) Improper normalization of named entities or temporal expressions during the claim extraction stage, making alignment difficult during verification (e.g., "last weekend" vs. an exact date). We hope to address these issues in future work.

\input{tables/poli_sno}

\input{tables/latestnews}

\section{Limitations}
News events are inherently dynamic, and the truth surrounding them can evolve over time. When verifying a news article being published in 2015 that discusses the average income increase ratio since 2001, it is crucial to obtain accurate data spanning from 2001 to 2015. This task presents challenges not only to search engines but also to LLMs. We have observed that our system performs more effectively when verifying more recent news articles. To close the gap, it requires truly understanding of timestamps by LLMs and the ability to accurately perform time sensitive calculations.

It has been noted that low-quality news articles frequently mix facts with opinions. In addition to verifying facts, it's important to distinguish the opinion segments within a news report. To accomplish this, it is crucial to integrate article-level linguistic features with retrieval-augmented fact verification methods.

Fake news can be deliberately created on a large scale. Beyond verifying individual articles, checking the authenticity of clusters of articles, can significantly enhance the detection effectiveness.

For practical considerations such as enhancing service robustness, reducing latency, and cutting costs, it is desirable to develop a smaller-sized LLM specifically for fake news detection. We plan to significantly invest in creating high-quality training data and explore advanced fine-tuning technologies to bridge the performance gap with GPT-4 in this area.

\section{Ethical Discussion}
Detecting fake news is a critical task with significant consequences. The effectiveness of this detection depends on various factors, such as the quality of searches, the impartial assessment of source credibility, and the language understanding capabilities of large language models (LLMs), among others. Our system aims to gather pertinent evidence from reputable sources, thereby aiding users in making informed decisions but not making those decisions for them. This approach is clearly outlined on our demo site.

\bibliography{anthology,custom}
\bibliographystyle{acl_natbib}

\appendix
\onecolumn

\section{Prompts}
\label{sec:appendix-prompt}

Here we list the prompts used in the pipeline: 
\input{tables/prompts_appendix}

\clearpage
\section{Sample Results}
\label{sec:appendix-sample}
We provide an illustration of the process involved in verifying a news article below. 
\input{tables/example}

\end{document}

%% file: figures/system.tex
\begin{figure*}[!ht]
    \centering
    
    \includegraphics[width=0.9\textwidth]{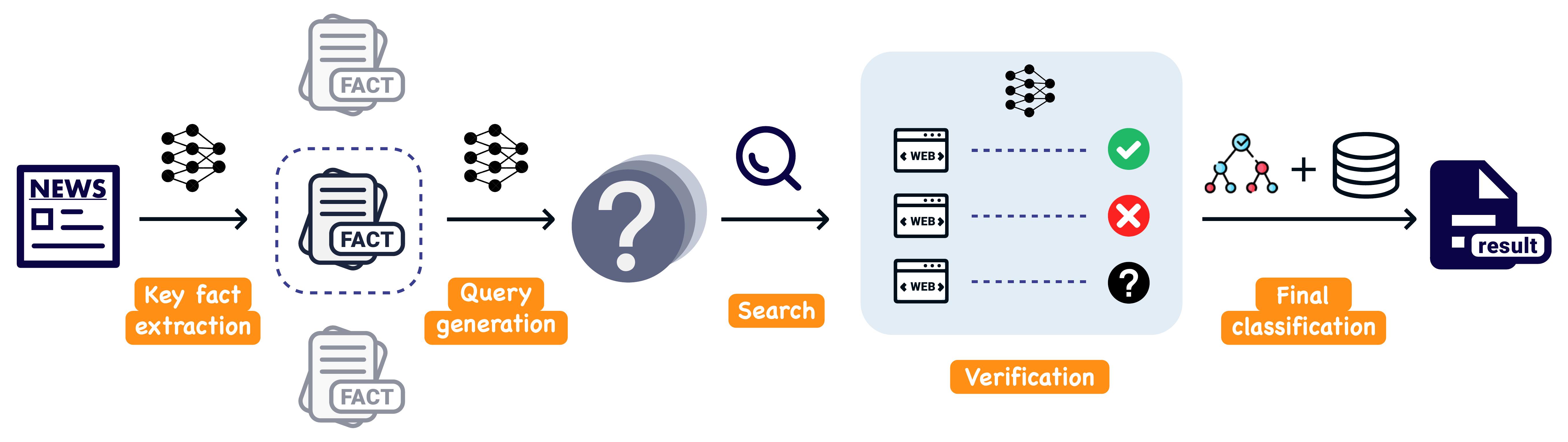}
    \caption{Main workflow of {\SystemName}. {\SystemName} includes the following steps: 1) extract key facts from the news to verify; 2) generate search queries for each extracted fact; 3) search; 4) verify the fact based on each search result; 5) aggregate all verifications with a final classification model. 
    }
    \label{fig:system}

\end{figure*}

%% file: tables/dataset.tex
\begin{table}[!t]
\centering
\scriptsize
\begin{tabular}{lcc}
\toprule
\textbf{Task/Dataset}                & \textbf{\# Samples} & \textbf{\% Samples}   \\
\midrule
Key Fact Extraction &  10299          &  18.52            \\
Query Generation    &  3000          &   5.39      \\
Fact Verification   &  23429          &  42.12        \\
QA with RAG                &  8091          &   14.55          \\
No robots\cite{no_robots}           &    9500        &    17.08        \\
WebGLM\cite{liu2023webglm} &  1300 &   2.34\\
\midrule
Total & 55619 & 100.0\\
\bottomrule
\end{tabular}
\caption{The distribution of the fine-tuning data from different tasks/datasets.}
\label{tab:dataset}
\end{table}

%% file: tables/key_task.tex
\begin{table}[!t]
\centering
\scriptsize
\begin{tabularx}{\linewidth}{Xccc}
\toprule
\textbf{Key Task}             & \textbf{ROUGE-1} & \textbf{ROUGE-2} & {\textbf{ROUGE-L}}         \\
\midrule
Key Fact Extraction & 0.678   & 0.497 & 0.655  \\
Query Generation    & 0.690   & 0.503 & 0.662 \\
Rationale Generation  & 0.637 & 0.449   & 0.600   \\
\bottomrule
\end{tabularx}
\caption{Performance of key tasks.}
\label{tab:result_keytask}
\end{table}

%% file: tables/buzzfeed.tex
\begin{table}[!t]
\centering
\scriptsize
\begin{tabularx}{\linewidth}{Xcccc}
\toprule
\textbf{Method}             & \textbf{Accuracy} & \textbf{Precision} & {\textbf{Recall}}   & {\textbf{F1}}       \\
\midrule
\citet{perez2017automatic} & 75.5   & 74.5 & 76.9 & 75.7   \\
\citet{shu2019beyond}         & 86.4          & 84.9   & 89.3     & 87.0      \\
\citet{zhou2020fakenewsearly}       & \textbf{87.9} & \textbf{85.7}  & \textbf{90.2} &  \textbf{87.9} \\
\midrule
Ours~(GPT)          &   79.1      &   81.2       &  75.8        &      78.4   \\
Ours~(Llama)      &  73.6       &  71.3      &  79.1        &   75.0  \\
\bottomrule
\end{tabularx}
\caption{Detection performance on BuzzFeedNews.}
\label{tab:result_buzzfeed}
\end{table}

%% file: tables/fakenewsnet.tex
\begin{table}[!t]
\centering
\scriptsize
\begin{tabular}{lp{0.3cm}p{0.52cm}p{0.43cm}p{0.4cm}p{0.52cm}p{0.42cm}p{0.39cm}}
\toprule
\textbf{Method} & \textbf{F1} & \textbf{F1-T} & \textbf{R-T} & \textbf{P-T} & \textbf{F1-F} & \textbf{R-F} & \textbf{P-F}  \\
\midrule
\citet{10.1145/3580305.3599873} & 72.9 & 75.7 & 78.0 & 73.5 & 70.2 & 68.1 & 72.8 \\
Ours~(GPT) & \textbf{80.3}   & \textbf{81.9}     & \textbf{85.9}    & \textbf{78.2}    & \textbf{78.3}     & \textbf{74.1}    &   \textbf{83.0}      \\
{Ours~(Llama)} & 77.3   & 79.0     & 82.3    &  75.9    & 75.3     & 71.9    &   79.1     \\
\bottomrule
\end{tabular}
\caption{Detection performance on Fakenewsnet.}
\label{tab:result_fakenewsnet}
\end{table}

%% file: tables/llm_generated.tex
\begin{table}[!t]
\centering
\scriptsize
\begin{tabular}{lcccc}
\toprule
\textbf{Dataset}    & \textbf{Written} & \textbf{Paraphrasing} & \textbf{Rewriting} & \textbf{Generating} \\
\midrule
\multicolumn{5}{l}{GPT-4-based Zero-shot Detector~(COT)~\citep{chen2023llmgenerated}}                           \\
\textbf{Politifact} & 62.6          & 56.0                  & 53.6                 & 41.6                  \\
\textbf{Gossipcop}  & 26.3          & 30.0                  & 25.0                 & 25.7                  \\
\textbf{CoAID}      & 81.0          & 82.2                  & 73.3                 & 52.7                  \\
\midrule
\multicolumn{5}{l}{Ours~(GPT)}                                                    \\
\textbf{Politifact} &   \textbf{63.7}           &   \textbf{62.2}                   &    \textbf{60.0}                  &      \textbf{60.7}                 \\
\textbf{Gossipcop}  &    \textbf{42.9}           &    \textbf{42.0}                   &    \textbf{40.3}                  &      \textbf{39.4}                 \\
\textbf{CoAID}      &   \textbf{83.7}            &    \textbf{86.0}                   &     \textbf{77.9}                 &      \textbf{69.8}                 \\
\midrule
\multicolumn{5}{l}{Ours~(Llama)}                                                     \\
\textbf{Politifact} &   56.3            &     55.9                  &     55.5                 &      51.1                 \\
\textbf{Gossipcop}  &    31.2           &    30.3                   &     34.6                 &       28.6                \\
\textbf{CoAID}      &    74.4           &     75.6                  &      70.9                &       60.5               \\
\bottomrule
\end{tabular}
\caption{Detection performance on LLMFake.}
\label{tab:result_llm_generated}
\end{table}

%% file: tables/poli_sno.tex
\begin{table}[!t]
\centering
\scriptsize
\begin{tabular}{llllllll}
\toprule
\textbf{Method} & \textbf{F1} & \textbf{F1-T} & \textbf{R-T} & \textbf{P-T} & \textbf{F1-F} & \textbf{R-F} & \textbf{P-F}  \\
\midrule
Ours~(GPT)       & 91.7   &  91.7    & 90.7    & 92.8    & 91.7     & 92.8    &   90.7      \\
Ours~(Llama)       & 85.6   &  85.9    &  86.4   &  85.3   &  85.3    &  84.8   &   85.9     \\
\bottomrule
\end{tabular}
\caption{Detection performance on PolitiFact-Snopes-2024.}
\label{tab:result_poli_sno}
\end{table}

%% file: tables/latestnews.tex
\begin{table}[!t]
\centering
\scriptsize
\begin{tabular}{lp{0.45cm}p{0.55cm}p{0.45cm}p{0.45cm}p{0.55cm}p{0.45cm}p{0.45cm}}
\toprule
\textbf{Method} & \textbf{F1} & \textbf{F1-T} & \textbf{R-T} & \textbf{P-T} & \textbf{F1-F} & \textbf{R-F} & \textbf{P-F}  \\
\midrule
Ours~(GPT)       & 89.9   &  87.6    &  84.8   & 90.7    &  91.5    &  93.7   & 89.4        \\
Ours~(Llama)       & 82.9   &  80.0    &  78.3   & 81.8    & 85.9     & 87.3    & 84.6       \\
\bottomrule
\end{tabular}
\caption{Detection performance on FakeNews2024.}
\label{tab:result_latest_news}
\end{table}

%% file: tables/prompts_appendix.tex
\begin{table}[h]
\small
\centering
\begin{tabular}{p{1\textwidth}}
    \toprule
    \textbf{\textsc{Main claim extraction}} \\
    \midrule
    Given the input content below, please summarize the single key claim. \\
    Input content: \{content\} \\
    Please output with the follow json format \{\{"key\_claim": XXX\}\}. \\
    Please output now: \\
    \midrule
    \textbf{\textsc{Key claims extraction}} \\
    \midrule
    Given the input content below, please extract distinct key claims. The key claims should be concrete enough containing clear context so that it can be efficiently verified. \\
    Input content: \{content\} \\
    Please output with the follow json format \{\{"key\_claims": [\{\{"claim": XXX\}\}, ...]\}\}.\\
    Please output now: \\
    \midrule
    \textbf{\textsc{Query generation}} \\
    \midrule
    Given the claim below, please generate a Google query which can be used to search content to verify this claim. \\
    Claim: \{claim\} \\
    Please output with the following JSON format \{\{"query": "XXX"\}\} \\
    Please output now: \\
    \midrule
    \textbf{\textsc{Content claim verification}} \\
    \midrule
    Below is one web search result \\
    Search Result: \\
    \{search\_result\} \\
    Below is a claim to be verified \\
    Claim: \{claim\} \\
    Please perform the following rules to generate an output with this json format : \{\{"support\_or\_negate\_or\_baseless": "support" or "negate" or "baseless", "confidence": "high" or "medium" or "low", "rationale": "XXX"\}\} \\
    Rule 1: if the search result content support the claim, set the "support\_or\_negate\_or\_baseless" field as "support", and offer a confident score and a rationale. \\
    Rule 2: if the search result content negate the claim, set the "support\_or\_negate\_or\_baseless" field as "negate", and offer a confident score and a rationale. \\
    Rule 3: if the search result content cannot either support or negate the claim, set the "support\_or\_negate\_or\_baseless" field as "baseless", and offer a confident score and a rationale. \\
    To clarify: if the content of the search results does not contradict the claim, but lacks some or all of the information presented in the claim, please use the label "baseless" rather than "negate". \\
    Please output now: \\
    \midrule
    \textbf{\textsc{Same news/relevant verification}} \\
    \midrule
    Below is one web search result. \\
    Search Result: \{search\_result\} \\
    Below is a claim: \\
    Claim:  \{claim\} \\
    Please make the following two investigations: \\
    1. Please check if the news article and the search result is about the same news story. \\
    2. Please check if the search result contains content (facts, opinions, or claims) related to the news article. \\
    Please output with the following json format : \\
    \{\{"about\_the\_same\_news\_story": "yes" or "no", "contains\_related\_content": "yes" or "no"\}\} \\
    Please output now: \\
    \bottomrule
\end{tabular}
\caption{Prompts used for key tasks.
}
\end{table}

%% file: tables/example.tex
\begin{table}[!h]
\renewcommand{\arraystretch}{1.2}
\small
\centering
    \begin{longtable}{p{1.0\textwidth}}
    \toprule
    \textbf{\textsc{News Article:}} Scientists Warn Eggs Are Causing Thousands of People to 'Suddenly' Form Blood Clots

In what appears to be another example of the global elite attempting to distract the public from the real cause of the surge in heart problems since the jab rollout, scientists now want us to believe that a nutrient found in eggs increases the risk of blood clotting.

Express.co.uk reports: The study conducted by Cleveland Clinic, suggested that choline could make the blood more prone to clotting.

Choline is sometimes sold in over-the-counter dietary supplements.

It is deemed an essential nutrient that is widely recommended for optimal health.

According to Medical News Today, it supports various bodily functions, including cellular growth and metabolism.\\






...
\\
    
    \midrule
    
    \textbf{\textsc{Claims:}} 1. A study conducted by Cleveland Clinic suggested that choline could make the blood more prone to clotting. \\
    2. Choline is deemed an essential nutrient that supports various bodily functions, including cellular growth and metabolism. \\
    3. Researchers found that consuming choline in high concentrations could lead to blood clotting. \\
    4. The interaction between choline and gut bacteria produces TMAO, which has been linked to an increased risk of blood clots, heart attack, and stroke.\\
    ...(8 claims extracted)\\
    \midrule
    \textbf{\textsc{Verify}}: A study conducted by Cleveland Clinic suggested that choline could make the blood more prone to clotting. \\
    \textbf{\textsc{Generated Queries:}} \\
    1. Cleveland Clinic study choline blood clotting  \\
    2. Researchers found that consuming choline in high concentrations could lead to blood clotting. \\
    \midrule
    \textbf{\textsc{Num of Search results:}} 18 \\
    \midrule
    \textbf{\textsc{Verification \& Rationale:}} \\
    Title: Study from 2017 misrepresented to link eggs to blood clots. \\
    Result: supported \\
    Domain: apnews.com \\
    Rationale: The search result explicitly mentions a study conducted by the Cleveland Clinic in 2017, which found that taking supplements in capsule form of choline 2014 a nutrient present in red meat, egg yolks, and other foods 2014 raised levels of a compound in the body (trimethylamine N-oxide, or TMAO) that could make the blood more prone to clotting. This directly supports the claim that consuming choline in high concentrations could lead to blood clotting.
    \begin{center}
    ---
    \end{center}
    Title: Cleveland Clinic Researchers First to Show Dietary Choline and Gut Bacteria Byproduct Linked with Increased Blood...\\
    Result: supported \\
    Domain: clevelandclinic.org\\
    Rationale: The search result explicitly states that Cleveland Clinic researchers have shown for the first time in humans that choline is directly linked to increased production of a gut bacteria byproduct that increases the risk of blood-clotting events like heart attack and stroke. This directly supports the claim that choline could make the blood more prone to clotting. \\
    \begin{center}
    ---
    \end{center}
    Title: Cleveland Clinic study did not conclude eggs are causing... \\
    Result: baseless \\
    Domain: reuters.com \\
    Rationale: The provided search result does not contain information about the effect of choline intake on blood clotting. It only mentions a study done by Cleveland Clinic regarding eggs, which is unrelated to the claim about choline. 

    \begin{center}
    ...
    \end{center}\\
    \midrule
    \textbf{\textsc{final decision:}} Supported\\
    \bottomrule
    \caption{An example of our pipeline ouput. Given that multiple claims can be extracted from a single article, we only exemplify the verification details of the first claim. The system generated two search queries related to the claim, resulting in the retrieval of 18 documents from the web. Based on the analysis of these documents, 14 documents are marked \textit{baseless} (irrelevant or not fully support the claim), whereas the remaining 4 documents \textit{support} the claim. By considering the sources credibility, the claim is classified as supported.}
    \end{longtable}
    
    \label{tab:example}
\end{table}